# Decoding Tourist Perception in Historic Urban Quarters with Multimodal Social Media Data: An AI-Based Framework and Evidence from Shanghai

Kaizhen Tan[1,2*], Yufan Wu[3]†, Yuxuan Liu[3]†, Haoran Zeng[3]†

[1] School of Economics and Management, Tongji University, Shanghai, China
[2] Heinz College of Information Systems and Public Policy, Carnegie Mellon University, Pittsburgh, PA, USA
[3] College of Architecture and Urban Planning, Tongji University, Shanghai, China

*Corresponding author. E-mail: kaizhent@cmu.edu; ORCID: 0009-0002-6459-578X
†These authors contributed equally to this work.

**Abstract.** Historic urban quarters are increasingly shaped by tourism and lifestyle consumption, yet planners often lack scalable evidence on what visitors notice, prefer, and criticize in these environments. This study proposes an AI-based, multimodal framework to decode tourist perception by combining visual attention, color-based aesthetic representation, and multidimensional satisfaction. We collect geotagged photos and review texts from a major Chinese platform and assemble a street view image set as a baseline for comparison across 12 historic urban quarters in Shanghai. We train a semantic segmentation model to quantify foregrounded visual elements in tourist-shared imagery, extract and compare color palettes between social media photos and street views, and apply a multi-task sentiment classifier to assess satisfaction across four experience dimensions that correspond to activity, physical setting, supporting services, and commercial offerings. Results show that tourist photos systematically foreground key streetscape elements and that the color composition represented on social media can differ from on-site street views, indicating a perception–reality gap that varies by quarter. The framework offers an interpretable and transferable approach to diagnose such gaps and to inform heritage management and visitor-oriented urban design.

## 1 Introduction

Historic urban quarters concentrate tangible and intangible heritage in a limited spatial envelope, carrying collective memory and place identity while supporting everyday life, cultural consumption, and tourism (Lynch, 1964; Assmann, 1995). As many cities reconfigure these areas through regeneration and destination branding, planners face a recurring tension: how to protect heritage character while maintaining experiential quality for visitors and residents. Understanding how tourists interpret historic quarters and evaluate their visual character is therefore central to human-centered planning in heritage contexts.

Tourist perception shapes destination image, which influences attractiveness, reputation, and long-term sustainability. It also feeds back into conservation priorities and planning strategies (Ginzarly et al., 2019). Tourism studies have long framed destination image and experience evaluation as a combined cognitive and affective construct, formed through direct encounters and through information before, during, and after travel (Echtner & Ritchie, 1993; Gartner, 1994; Baloglu & McCleary, 1999; Beerli & Martín, 2004). In heritage tourism, perception is additionally conditioned by authenticity judgments and by whether visitors interpret a place as connected to their own heritage, which shapes evaluation and behavioral intentions (MacCannell, 1973; Poria et al., 2003). Classic sociological accounts further emphasize that tourism involves a socially mediated gaze that structures what visitors attend to, photograph, and later remember (Urry, 1990). These perspectives suggest that tourist perception is not a single outcome, but a linked process that spans attention, representation, and appraisal.

Conventional studies of tourist experience in historic settings have often relied on surveys, interviews, and in-situ observation. These approaches provide rich insights but scale poorly and can be affected by reporting delay and respondent subjectivity. With the growth of user-generated content (UGC), researchers can observe tourism experience at scale through geo-tagged photos and accompanying texts posted on platforms such as Flickr and Instagram (Zhang et al., 2019; Kang et al., 2021; Cho et al., 2022). Reviews of photo-based tourism research show rapid expansion, yet the literature tends to cluster into separate streams, such as photo-sharing motivation, destination image inference, spatiotemporal behavior, or impacts, with limited integration across data types and analytical goals (Li et al., 2023). Urban informatics research also stresses that UGC is informative but uneven, with platform and demographic biases that must be considered when interpreting results and generalizing findings (Gao et al., 2021). Related validation work indicates that geo-tagged photo traces can proxy visitation patterns in certain contexts, while also documenting representativeness issues and spatial sampling biases (Wood et al., 2013; Tenkanen et al., 2017).

A central challenge in historic quarters is that visitors often evaluate not only functional conditions, but also atmosphere, coherence, and perceived heritage character. In this study, we use the term urban aesthetics to refer to evaluative responses to urban scenes that arise from perceptual cues interpreted through cultural meaning, rather than treating aesthetics as a purely formal property of colors or shapes. Environmental and design research suggests that building exteriors are judged through multiple types of variables, including formal qualities and symbolic meaning, mediated by learned schemas and cultural frames (Nasar, 1994). Recent work further argues that urban color is a culturally embedded resource whose "local" character can be weakened by homogenizing global aesthetics, making color governance relevant to place identity and urban sustainability (Xue et al., 2025). In this sense, chromatic composition is a measurable perceptual cue that may signal ambience or heritage character, but it does not exhaust aesthetic experience. This framing clarifies what a color-based analysis can support and what it should not claim.

These strands of scholarship motivate an integrated analytic chain that connects what tourists notice, what they curate in representation, and how they evaluate their experience. Visual attention relates to what is foregrounded in shared photos, reflecting selective capture shaped by the tourist gaze (Urry, 1990). Representation can be operationalized by comparing shared photos with on-site visual baselines, which helps separate place attributes from platform-mediated curation. Appraisal is expressed in review texts that evaluate experience components. Because experience evaluations are shaped by both the physical setting and the service environment, linking image-based indicators with dimension-specific sentiment can connect what is present on-site with what is evaluated in narration (Bitner, 1992).

Despite substantial progress, three gaps remain for multimodal analysis of tourist perception in historic urban quarters. First, many UGC studies analyze visual content and review texts in isolation, which limits the ability to connect what is visually foregrounded with what is later praised or criticized in narratives (Li et al., 2023). Second, computational studies of urban imagery often focus on façade harmony or other formal metrics without clarifying how measurable cues relate to evaluative judgments in heritage contexts (Nasar, 1994; Zhou et al., 2022; Yang et al., 2024). Third, historic quarters as bounded cultural spaces remain relatively underexamined compared with broader destination studies, and existing multimodal case work often emphasizes recognition of heritage elements rather than aesthetic expectation and satisfaction mechanisms (Bai et al., 2023). Together, these gaps call for a framework that links attention, representation, and appraisal in a single analytical chain and that provides interpretable indicators for planning practice.

To move beyond a purely technical description of urban color, it is necessary to examine the gap between a quarter's on-site visual output and tourists' aesthetic expectations as expressed through selective representation. We address this by comparing the chromatic composition of tourist-shared photos with street-view baselines, which helps identify potential mismatches between representation and on-site scenes. We then relate these signals to review-based satisfaction patterns to support planning-relevant diagnosis.

We propose a multidimensional AI-powered framework to decode tourist perception in historic urban quarters, with an exploratory case study in Shanghai. As illustrated in Figure 1, the framework integrates three modules: visual focus detection, aesthetic preference analysis through perception–baseline comparison, and satisfaction assessment from textual reviews. Conceptually, tourist perception is modeled as a linked process that moves from attention to representation and then to appraisal.

Empirically, we aim to generate planning-relevant indicators that connect what is present, what is portrayed, and what is evaluated.

This study makes three contributions. First, it develops an integrated multimodal framework that explicitly links visual attention, representation, and appraisal, supporting planning decisions in historic quarters. Second, it provides an application-oriented semantic taxonomy and a semantic segmentation pipeline tailored to tourist-shared imagery in historic street environments, enabling fine-grained measurement of what visitors visually foreground. Third, it introduces a perception–reality comparison of chromatic composition by aligning social media photos with street-view baselines, offering an interpretable way to diagnose aesthetic mismatch and relate it to dimension-specific satisfaction signals.

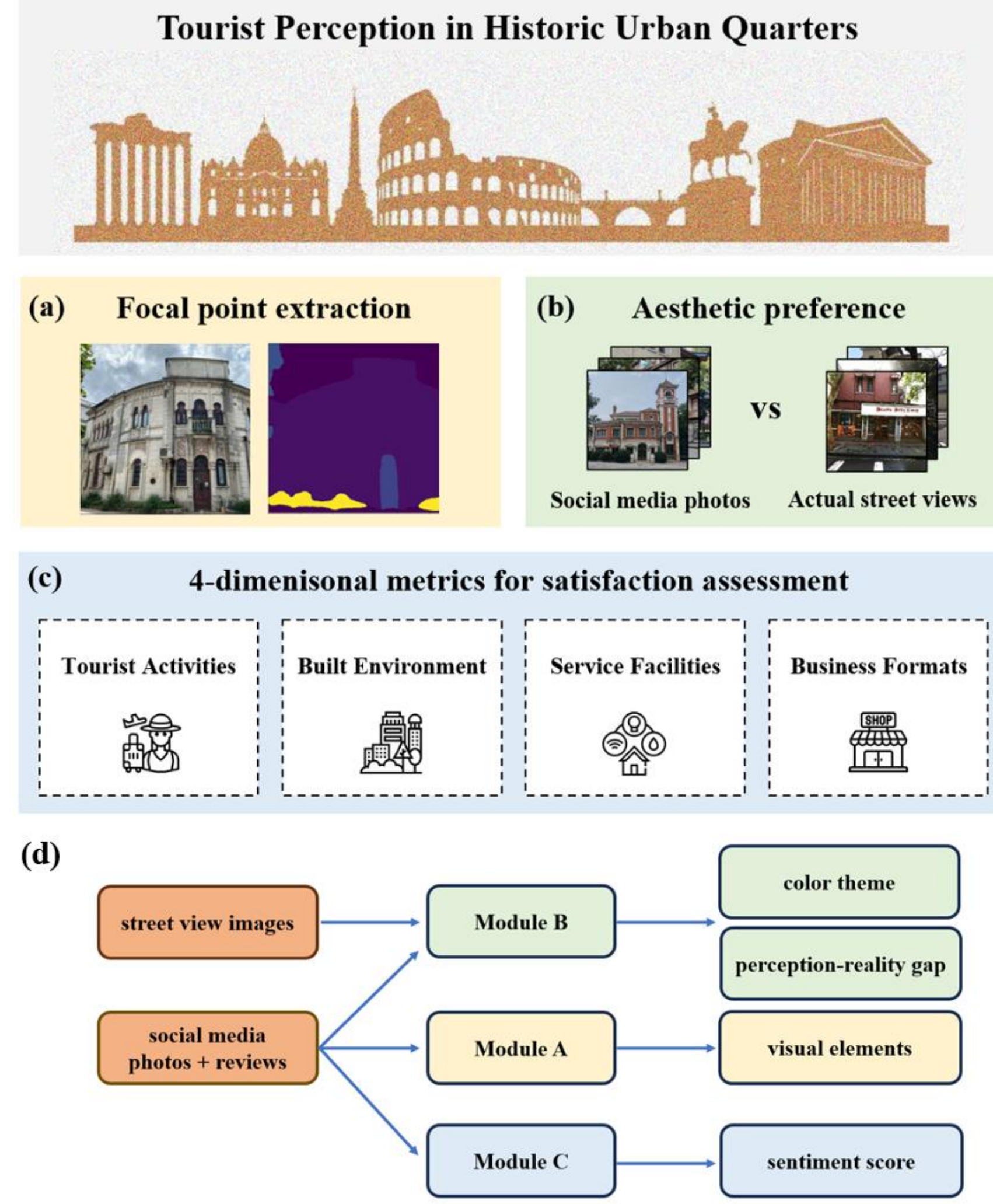

**Fig. 1** Conceptual overview of the AI-powered framework **(a)** Semantic segmentation identifies visual focus areas from tourist photos **(b)** Comparative analysis of dominant color composition between social media photos and street-view baselines reveals signals of aesthetic expectation; **(c)** Satisfaction is operationalized with four planning-relevant dimensions that capture recurring components in tourism experience and service environment research, including visitor activities, perceived built setting, supporting service facilities, and the commercial environment (Bitner, 1992; Echtner & Ritchie, 1993; Poria et al., 2003; Ashworth & Tunbridge, 2000) **(d)** The flowchart summarizes the interconnections among inputs, modules, and outputs

## 2 Methodology

### 2.1 Case Study

Shanghai is selected as the case study area. In this study, historical and cultural blocks designated by the Shanghai Municipal Government refer to urban areas where historic buildings are concentrated, and where architectural style, spatial form, and streetscape together represent the cultural character of a specific historical period in Shanghai. Figure 2(a) situates the study area and delineates Shanghai's central urban area, while Figure 2(b) maps the 12 designated historical and cultural blocks within it: People's Square, West Nanjing Road, The Bund, Shanyin Road, Yuyuan Road, Tilanqiao, Xinhua Road, Jiangwan, Laochengxiang, Hongqiao Road, Hengshan Road–Fuxing Road, and Longhua Road. These areas integrate distinctive styles from different stages of Shanghai's urban development, and reflect the city's modern achievements and evolution in terms of economy, culture, and everyday life.

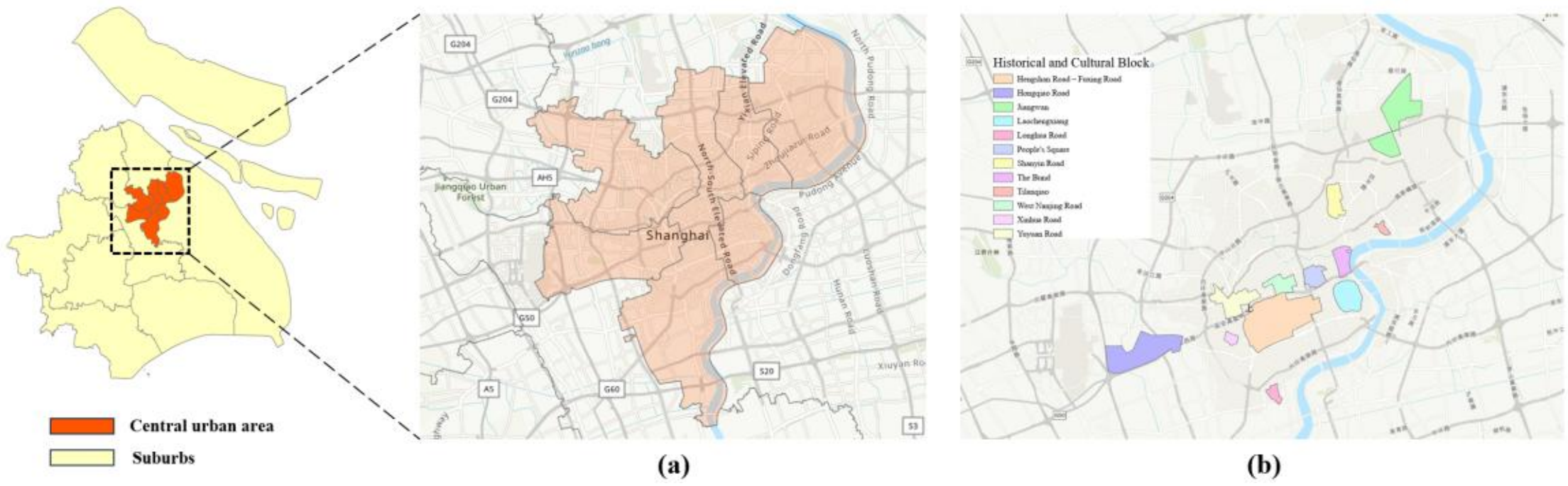

**Fig. 2** Study area in Shanghai (a) Boundary of the central urban area (b) 12 historical and cultural blocks

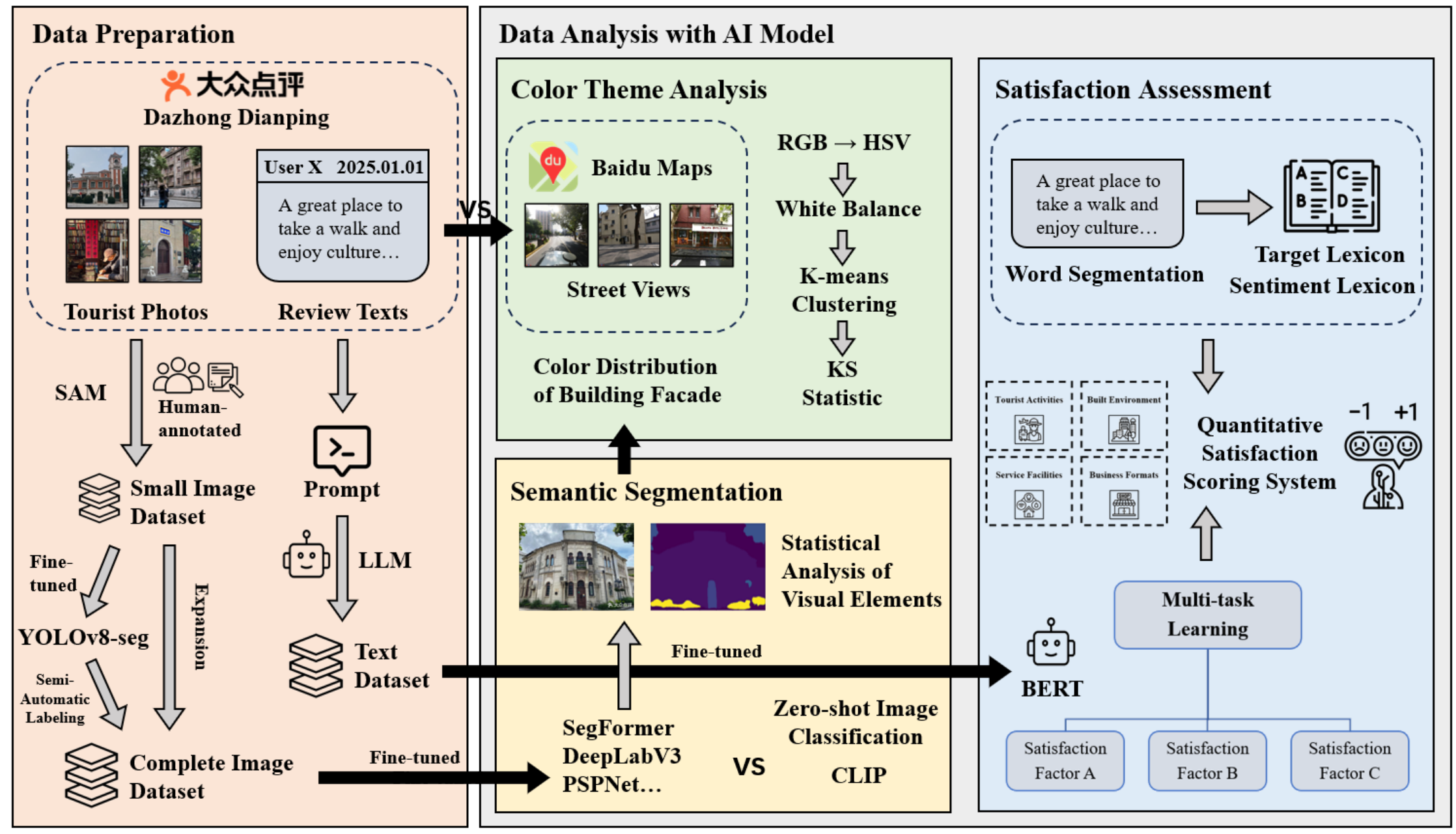

**Fig. 3** Technical roadmap of the AI-powered framework

## 2.2 Data Preparation

The overall workflow of the proposed framework is illustrated in Figure 3, which outlines the key steps of data collection, preprocessing, model training, and perceptual analysis. We collected tourist photos and review texts related to the selected historic urban quarters from the Dianping mobile app, and retrieved street view images from Baidu Maps as a real-world baseline for the color distribution comparison. Dianping data were crawled in late October 2024, covering reviews posted between 20 September 2018 and 28 October 2024. The raw crawl contains 33,168 reviews. After deduplication based on identical review text and posting time, 32,949 unique reviews remained. Among them, 28,649 reviews include at least one photo, yielding 168,137 photo files in total (mean 5.9 photos per photo-containing review). To reduce imbalance across quarters, we constructed a balanced photo subset by selecting one representative photo per photo-containing review and limiting each quarter to at most 5,000 photo posts, resulting in 20,968 photos for the image-based modules.

For each historic quarter, we first identified representative points of interest on Dianping (e.g., street segments, landmark buildings, or designated tourist spots) and used their listing pages as entry points for crawling. For each review, we stored the shop URL, shop name and address, posting time, star rating, full text, and all associated photo URLs. We removed 219 duplicate review records detected by exact matching of review text and posting time before downstream analysis.

Street view images were retrieved by sampling points every 20 meters along the street network within each quarter. At each sampling point, four views were requested with headings of 0°, 90°, 180°, and 270° (pitch = 0). For temporal alignment, we recommend recording the capture year provided by the street view service (when available) and selecting panoramas closest to the median year of social media posts in each quarter.

To construct a semantic segmentation dataset, the pre-trained Segment Anything Model (SAM) (Kirillov et al., 2023) was used as an auxiliary tool to assist with manual annotation. A high-quality training set was created by manually labeling 1,000 tourist photos. A Yolov8-Seg model (Jocher et al., 2023) was then fine-tuned on this initial set to develop an automatic labeling model. This model was subsequently integrated into the annotation software to accelerate the labeling process, enabling the expansion of the training set and the construction of a complete dataset. Initially, 26 categories were defined based on preliminary discussions, aiming to cover the main visual elements present in social media imagery. However, during the annotation process, some categories proved difficult to distinguish or appeared too infrequently in the dataset. Based on descriptive statistics and practical applicability, the final taxonomy was refined to 22 categories. Definitions of these categories are provided in the appendix.

Before constructing the multi-dimensional sentiment scoring dataset for deep learning, a small dataset was built using rule-based named entity recognition and sentiment word matching methods to perform the same task on a limited sample. The four dimensions of satisfaction (Tourist Activities, Built Environment, Service Facilities, and Business Formats) were determined based on prior survey findings, reflecting commonly held visitor concerns. We used the DeepSeek API to generate sentiment labels automatically by leveraging large language model capabilities. For each dimension, sentiment was labeled as {-1, 0, 1}, where -1 indicates dissatisfaction, 1 indicates satisfaction, and 0 denotes either no mention or a neutral attitude. Prompt design details are included in the appendix.

### 2.3 Semantic Segmentation

No single model has been universally recognized as optimal for semantic segmentation of tourist photographs. Model performance largely depends on dataset characteristics and application contexts. Even minor differences between datasets can lead to noticeable variation in high-dimensional feature mapping. Therefore, several state-of-the-art pre-trained models were evaluated, including SegFormer (Xie et al., 2021), Mask2Former (Cheng et al., 2022), DeepLabV3 (Chen et al., 2018), and PSPNet (Zhao et al., 2017). Fine-tuning experiments showed that Deeplabv3+MobilenetV2 (Sandler et al., 2018) and Mask2Former+Swin-L (Liu et al., 2021) performed well during the first 100 epochs. Considering the trade-offs between parameter size, computational cost, and inference speed, we ultimately adopted Deeplabv3 as the segmentation framework and MobilenetV2 as the backbone network.

Descriptive statistics of the labeled data revealed a significant class imbalance, with certain categories (e.g., fountain, animal, vendor) having substantially fewer samples, which resulted in poor recognition performance for these classes. To address this, class weights were incorporated into the loss function, assigning higher weights to underrepresented categories to increase their training emphasis. The weighting strategy was informed by statistical analysis and inspired by the application of Focal Loss in long-tailed data distributions, aiming to improve small-sample class recognition.

To optimize memory efficiency and prevent overfitting due to excessive parameters, a two-stage training strategy was adopted. In the freezing stage, higher-level features were trained while freezing lower layers to reduce GPU memory usage. In the unfreezing stage, all layers were fine-tuned jointly to maximize model performance. Finally, the trained segmentation model was applied to tourist photos collected from Dianping, generating statistics on the number and pixel proportions of each category. Results were compared with visual feature detection outputs from the CLIP model (Radford et al., 2021).

### 2.4 Color Theme Analysis

This study applies color space transformation and clustering algorithms to extract dominant color themes from images. These colors are classified based on the Chinese color system (Huang et al., 2007) to analyze spatial patterns and variations in color distribution across different quarters. Images were first preprocessed by converting from RGB to HSV color space, which better reflects human color perception. The Gray World algorithm was used for white balance correction to mitigate lighting effects and ensure consistency in color representation. K-means clustering was applied to the transformed color data to identify dominant colors in each image. The pixel count for each cluster center was computed, and the dominant colors were then categorized according to the Chinese color system. For each image, a top-5 color palette and corresponding textual data were generated, along with a top-20 color summary. Color distribution curves were fitted, and inter-quarter differences were evaluated using the Kolmogorov-Smirnov (K-S) divergence. Finally, semantic segmentation was applied to both tourist photos and actual street views to compare the distribution of façade colors. This allowed for a detailed analysis of aesthetic alignment or mismatch between perceived and real-world visual elements.

### 2.5 Satisfaction Assessment

We operationalize tourist satisfaction using four dimensions that align with how visitors typically narrate experiences and how planners can act on feedback. Tourist Activities capture experience content and participation; Built Environment captures perceived streetscape and heritage qualities; Service Facilities capture access, comfort, and supporting amenities; and Business Formats capture the influence of retail and consumption-oriented functions. This structure improves interpretability and allows results to be translated into actionable design and management interventions.

We adopt both traditional lexicon-based methods and a deep learning model. The traditional method involves customized Chinese word segmentation, a target lexicon, and a sentiment lexicon. We develop a quantitative satisfaction scoring scheme in which adjectives and evaluative terms are assigned weighted scores, adjusted by negations and degree adverbs. To ensure dimension-specific relevance, we segment reviews into clauses using fuzzy keyword matching and retain only sub-sentences that contain both target terms and sentiment cues, rather than applying general-purpose tokenization to full texts.

In parallel, we train a multi-task Bidirectional Encoder Representations from Transformers (BERT) model (Devlin et al., 2019) on labeled review texts to predict sentiment polarity in each dimension. We use BERT-base-Chinese as the backbone and attach four independent classification heads to output three-class sentiment labels (positive, neutral, negative) for the four dimensions.

## 3 Results

### 3.1 Focal Point Extraction

After training the semantic segmentation model, we applied it to the tourist photos to analyze the dominant visual elements in social media imagery. The results demonstrate that the model performs well in identifying core object categories, showing generalization across scenes with different primary subjects such as people and buildings. Figure 4 presents a sample of segmentation results. To evaluate the overall distribution of visual elements, we selected photos from 12 typical historic urban quarters and calculated the total number of pixels and corresponding proportions for each semantic class, excluding the background category. The aggregated results are shown in Figure 5.

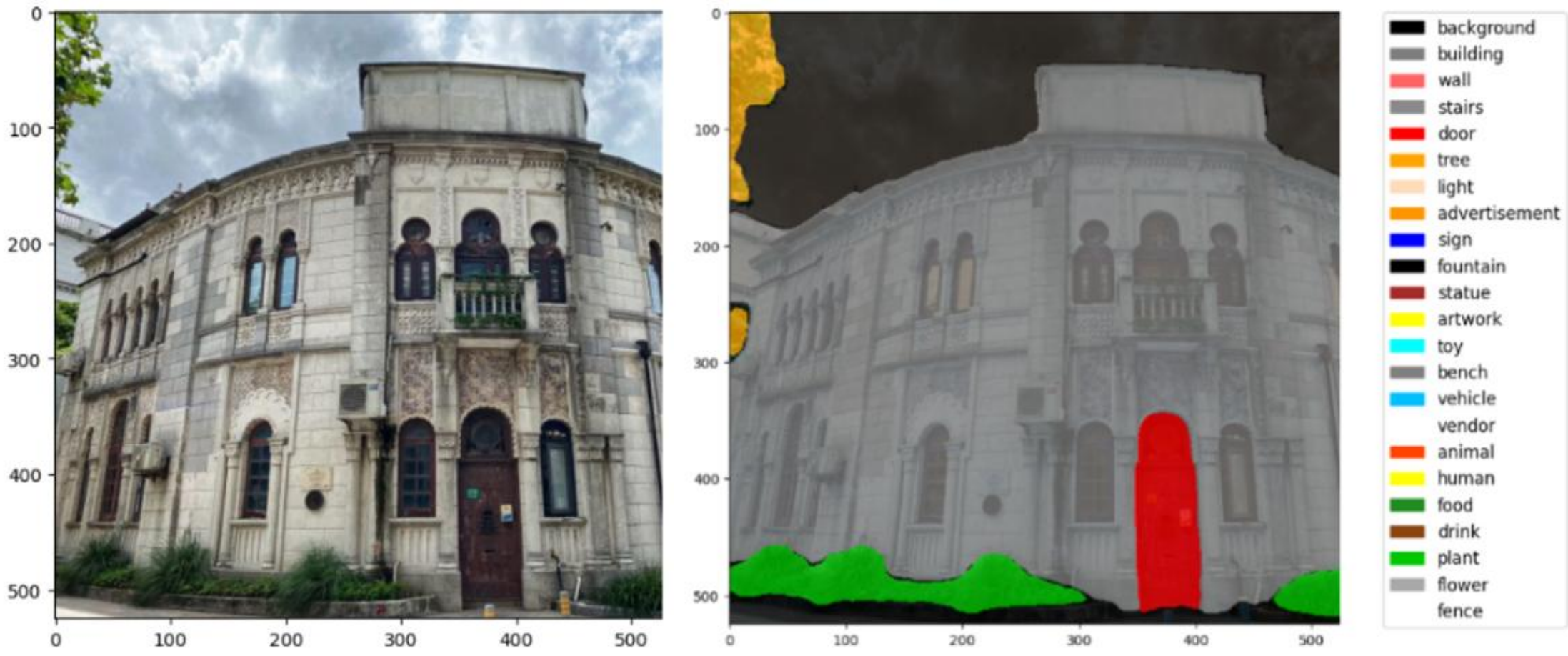

**Fig. 4** A sample of segmentation results

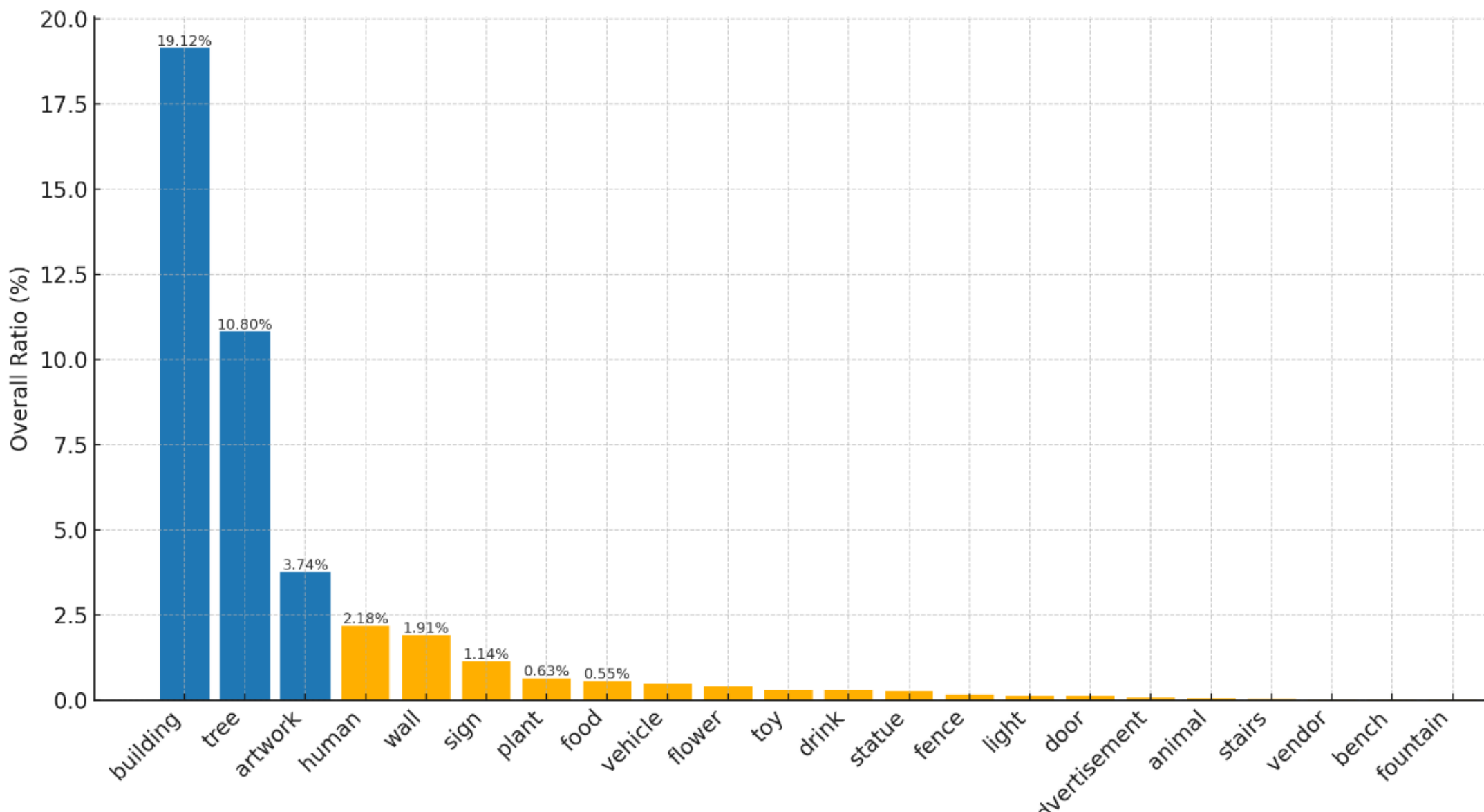

**Fig. 5** Overall distribution of semantic classes

Overall, the distribution aligns with intuitive understanding and reflects the composition of real-world streetscapes in historic quarters. Buildings account for the largest share, comprising 19.12% of total pixels, followed by trees at 10.80% and artworks at 3.74%. Other relatively prominent categories include humans, walls, and signs, all exceeding 1%. Elements such as advertisements, street vendors, and fountains occupy minimal proportions, indicating their lower frequency in typical street scenes. Figure 6 presents a heatmap showing the semantic class ratios for each of the 12 quarters. For every quarter, we list the categories with pixel proportions exceeding 1%. Analysis of these distributions in relation to actual urban features reveals several patterns. In leisure-oriented commercial areas such as Hengshan Road, Huaihai Road, Tian'ai Road, and Xinhua Road, greenery and historic buildings dominate the visual composition. These quarters are typically characterized by a high concentration of trees and architectural heritage, which environmental psychology and urban design research associates with higher preference and perceived comfort, thereby reinforcing both everyday livability and perceived cultural value (Smardon, 1988; Nasar, 1994).

In bustling commercial areas like Huaihai Road and Wukang Road, vehicles appear more frequently, reflecting the concentration of traffic in these zones. For example, Wukang Mansion, a popular photo spot, is often photographed from across the street, resulting in vehicles being captured within the frame and thus increasing the visual share of the vehicle category.

In high-traffic zones such as the Bund and Wujiang Road, humans rather than trees form the second most prominent visual element. This suggests distinct patterns of urban use. The Bund is dominated by waterfront cityscapes with limited greenery, making buildings and people the primary elements. Wujiang Road, known for its food culture, also shows a notably higher proportion of food-related imagery, reflecting its lively street life and commercial appeal.

Certain quarters also display strong artistic character. In particular, Tian'ai Road, Yuyuan Road, and Hengshan Road have significantly higher proportions of artworks compared to other areas. This reflects their role as cultural and creative hubs within the city. For instance, the graffiti wall on Tian'ai Road accounts for an artwork ratio as high as 12.81%. Similarly, Yuyuan Road and Hengshan Road are associated with cultural industries and public art installations. Yuyuan Garden and Shanyin Road also exhibit considerable artistic features, emphasizing their unique cultural atmosphere.

These findings suggest that the semantic segmentation model can adapt well to different street environments and effectively extract core visual elements. The statistical results align closely with the spatial characteristics of the quarters. For example, the heavy pedestrian flows at the Bund and Wujiang Road, the artistic prominence of Tian'ai Road, and the vehicular presence in commercial zones are all clearly reflected in the data. This method of analysis provides valuable insight for urban planning and commercial decision-making. Metrics such as greenery ratio can inform assessments of urban livability, while the proportion of food-related content may serve as a proxy for culinary attractiveness.

While the segmentation-based approach provides fine-grained quantification of visual elements, it has certain limitations. It does not account for overall scene semantics or the relative importance of different elements. To address this, we used the CLIP (Contrastive Language-Image Pre-Training) model, which identifies high-level semantic concepts by matching images with textual labels. Unlike pixel-based segmentation, CLIP offers a conceptual understanding of urban imagery. Although it may miss fine details, it complements segmentation by revealing the dominant meanings conveyed in tourist photos. The distribution of categories detected by CLIP is presented in the appendix.

| Semantic Class | Duolun Road | Hengshan Road | Huaihai Road | West Nanjing Road | Shanyin Road | Tian'ai Road | The Bund | Wujiang Road | Wukang Road | Xinhua Road | Yuyuan Road | Yuyuan Garden |
|---|---|---|---|---|---|---|---|---|---|---|---|---|
| building | 30.85 | 14.32 | 15.01 | 21.01 | 29.26 | 12.87 | 23.24 | 12.94 | 25.62 | 14.89 | 11.81 | 14.63 |
| tree | 15.21 | 19.38 | 15.93 | 7.26 | 15.67 | 15.29 | 2.88 | 0.00 | 15.01 | 25.10 | 10.18 | 12.91 |
| sign | 2.31 | 5.18 | 0.00 | 1.11 | 3.09 | 1.24 | 0.00 | 0.00 | 0.00 | 1.91 | 1.36 | 0.00 |
| wall | 1.90 | 0.00 | 0.00 | 0.00 | 3.65 | 7.25 | 0.00 | 0.00 | 1.45 | 1.89 | 2.41 | 0.00 |
| plant | 1.62 | 1.65 | 0.00 | 0.00 | 0.00 | 0.00 | 0.00 | 0.00 | 0.00 | 0.00 | 0.00 | 0.00 |
| statue | 1.36 | 0.00 | 0.00 | 0.00 | 0.00 | 0.00 | 0.00 | 0.00 | 0.00 | 0.00 | 0.00 | 0.00 |
| artwork | 1.19 | 3.03 | 1.00 | 1.15 | 1.95 | 12.81 | 0.00 | 1.12 | 0.00 | 1.05 | 3.39 | 2.50 |
| flower | 0.00 | 1.05 | 0.00 | 1.97 | 0.00 | 0.00 | 0.00 | 0.00 | 0.00 | 0.00 | 0.00 | 0.00 |
| food | 0.00 | 0.00 | 2.87 | 1.00 | 0.00 | 0.00 | 0.00 | 2.46 | 0.00 | 0.00 | 0.00 | 0.00 |
| human | 0.00 | 0.00 | 2.59 | 2.47 | 0.00 | 1.54 | 3.18 | 4.48 | 2.18 | 1.26 | 2.09 | 2.10 |
| vehicle | 0.00 | 0.00 | 1.05 | 0.00 | 0.00 | 0.00 | 0.00 | 0.00 | 1.08 | 0.00 | 0.00 | 0.00 |
| toy | 0.00 | 0.00 | 0.00 | 0.00 | 0.00 | 0.00 | 0.00 | 0.00 | 0.00 | 0.00 | 1.36 | 0.00 |
| drink | 0.00 | 0.00 | 0.00 | 0.00 | 0.00 | 0.00 | 0.00 | 0.00 | 0.00 | 0.00 | 0.00 | 1.63 |

**Fig. 6** Heatmap of semantic class ratios across 12 historic urban quarters

## 3.2 Aesthetic Preference: Color Composition and Expectations

For each tourist photo, we extracted five dominant colors using K-means clustering and visualized them as proportional color blocks. As this method is clustering-based, some color merging and representation errors may occur. In addition, we categorized all pixels into one of 1,000 predefined hues in the Chinese color system and selected the top 20 most frequent colors. While this classification method is less prone to error, it does not consider color grouping and focuses solely on pixel-level frequency. A sample result is shown in Figure 7, where the original image is compared with the K-means color palette and the top 20 classified colors.

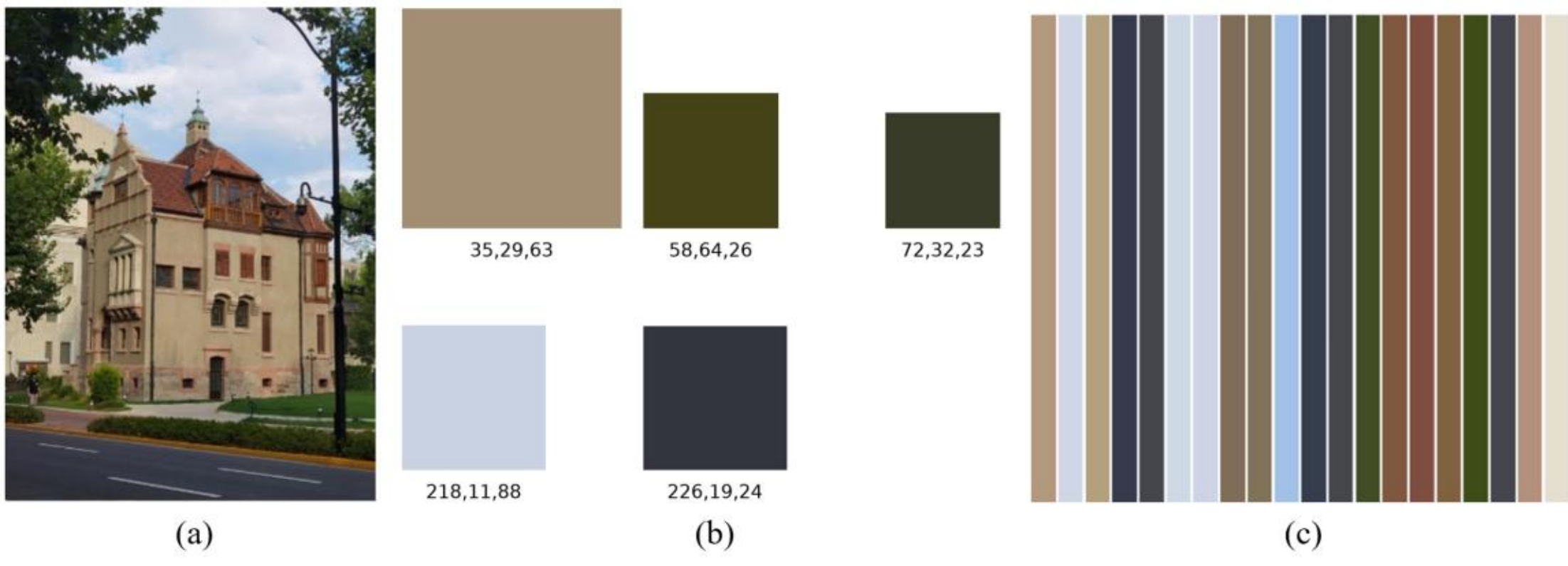

**Fig. 7** Example of color extraction results **(a)** Original photo **(b)** K-means clustered top-5 dominant colors **(c)** Top-20 dominant colors using Chinese color system classification

At the quarter level, we analyzed spatial patterns of color composition by examining the distribution of hue (H) values. Figure 8 displays the overall hue distribution across historic quarters along with fitted curves. Both the aggregate and quarter-level distributions show consistent bimodal patterns, with peaks around the orange-red (H ≈ 0–30) and blue-violet (H ≈ 100–120) ranges, indicating that these tones are particularly prevalent in the urban landscape.

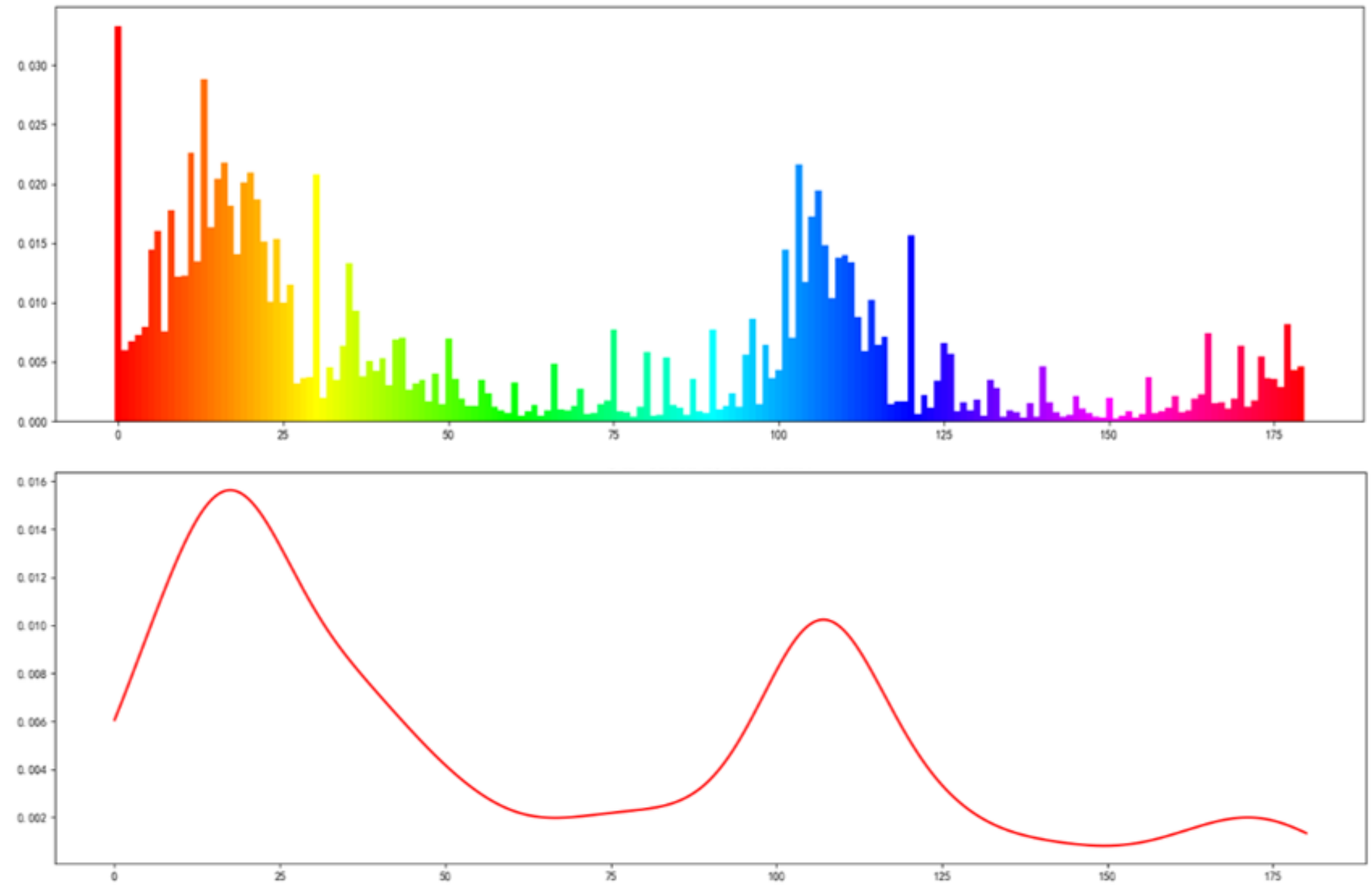

**Fig. 8** Overall hue (H) distribution of tourist photos across historic urban quarters, with fitted curve

To quantify inter-quarter differences in color composition, we applied the Kolmogorov–Smirnov (KS) divergence metric. Figure 9 presents a KS divergence matrix of hue distributions across 12 historic urban quarters. A smaller KS value indicates greater similarity in color distribution. Figure 10 compares the hue distributions of Huaihai Road and the Bund. The results show that Fuxing Road and Huaihai Road, both located within the Hengfu historical block, exhibit significant color distribution differences when compared to the Bund, with KS divergence values reaching 0.244 and 0.247, respectively. This divergence may result from the functional characteristics. Quarters characterized by enclosed indoor spaces, such as cultural shops or food streets, tend to exhibit warmer tones (e.g., red, orange, yellow). In contrast, open outdoor areas such as waterfront promenades or historical plazas are dominated by cooler hues (e.g., blue, grey, green). The Bund, situated along the Huangpu River, shows a higher proportion of blue tones influenced by the sky and water bodies, which contributes to its distinctly cooler visual profile.

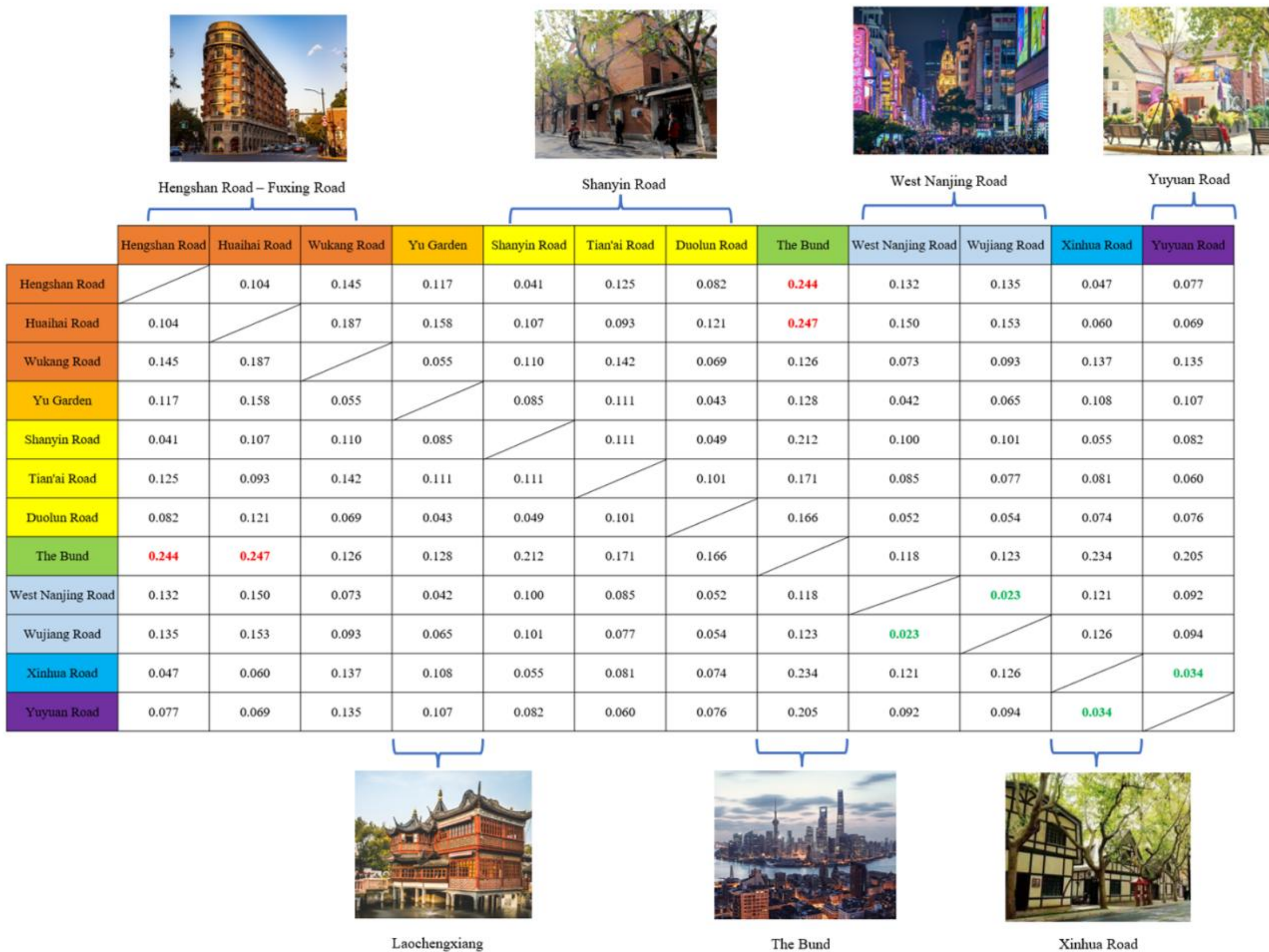

| | Hengshan Road | Huaihai Road | Wukang Road | Yu Garden | Shanyin Road | Tian'ai Road | Duolun Road | The Bund | West Nanjing Road | Wujiang Road | Xinhua Road | Yuyuan Road |
|---|---|---|---|---|---|---|---|---|---|---|---|---|
| Hengshan Road | | 0.104 | 0.145 | 0.117 | 0.041 | 0.125 | 0.082 | 0.244 | 0.132 | 0.135 | 0.047 | 0.077 |
| Huaihai Road | 0.104 | | 0.187 | 0.158 | 0.107 | 0.093 | 0.121 | 0.247 | 0.150 | 0.153 | 0.060 | 0.069 |
| Wukang Road | 0.145 | 0.187 | | 0.055 | 0.110 | 0.142 | 0.069 | 0.126 | 0.073 | 0.093 | 0.137 | 0.135 |
| Yu Garden | 0.117 | 0.158 | 0.055 | | 0.085 | 0.111 | 0.043 | 0.128 | 0.042 | 0.065 | 0.108 | 0.107 |
| Shanyin Road | 0.041 | 0.107 | 0.110 | 0.085 | | 0.111 | 0.049 | 0.212 | 0.100 | 0.101 | 0.055 | 0.082 |
| Tian'ai Road | 0.125 | 0.093 | 0.142 | 0.111 | 0.111 | | 0.101 | 0.171 | 0.085 | 0.077 | 0.081 | 0.060 |
| Duolun Road | 0.082 | 0.121 | 0.069 | 0.043 | 0.049 | 0.101 | | 0.166 | 0.052 | 0.054 | 0.074 | 0.076 |
| The Bund | 0.244 | 0.247 | 0.126 | 0.128 | 0.212 | 0.171 | 0.166 | | 0.118 | 0.123 | 0.234 | 0.205 |
| West Nanjing Road | 0.132 | 0.150 | 0.073 | 0.042 | 0.100 | 0.085 | 0.052 | 0.118 | | 0.023 | 0.121 | 0.092 |
| Wujiang Road | 0.135 | 0.153 | 0.093 | 0.065 | 0.101 | 0.077 | 0.054 | 0.123 | 0.023 | | 0.126 | 0.094 |
| Xinhua Road | 0.047 | 0.060 | 0.137 | 0.108 | 0.055 | 0.081 | 0.074 | 0.234 | 0.121 | 0.126 | | 0.034 |
| Yuyuan Road | 0.077 | 0.069 | 0.135 | 0.107 | 0.082 | 0.060 | 0.076 | 0.205 | 0.092 | 0.094 | 0.034 | |

**Fig. 9** KS divergence matrix of hue distributions across 12 historic urban quarters (quarters with similar color distributions are typically part of the same historical and cultural block)

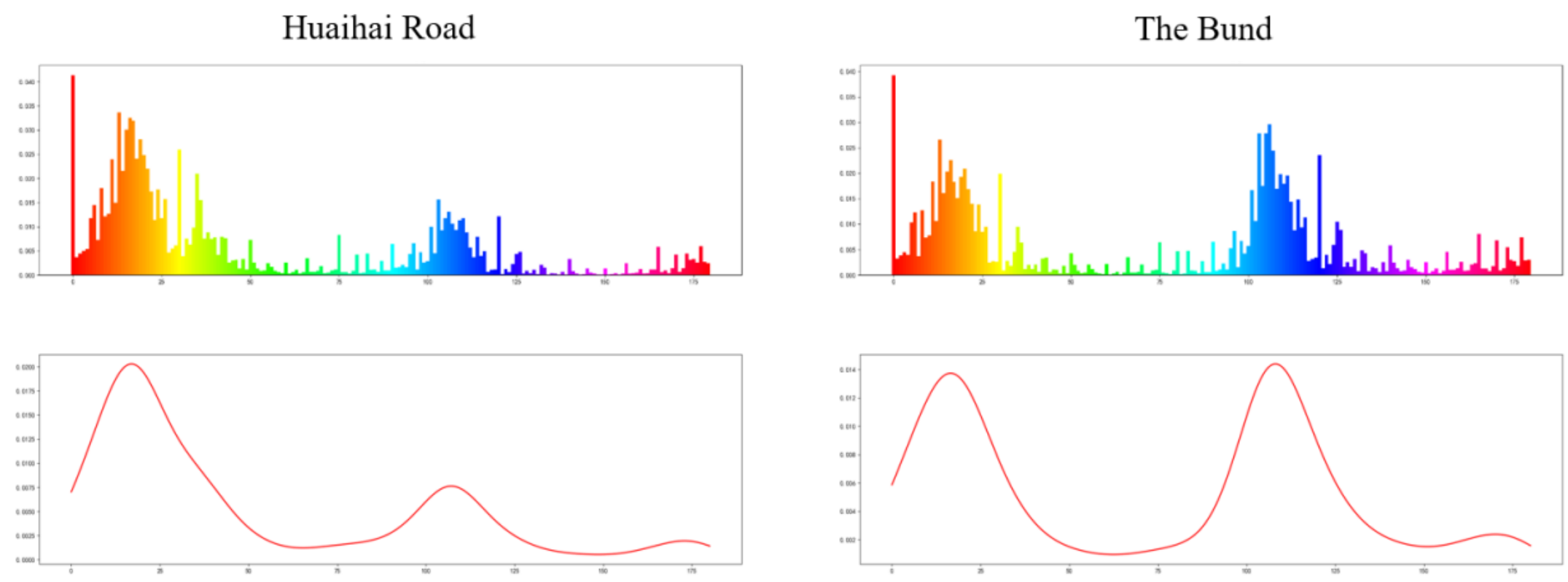

**Fig. 10** Comparison of hue (H) distributions and fitted curves for Huaihai Road and the Bund

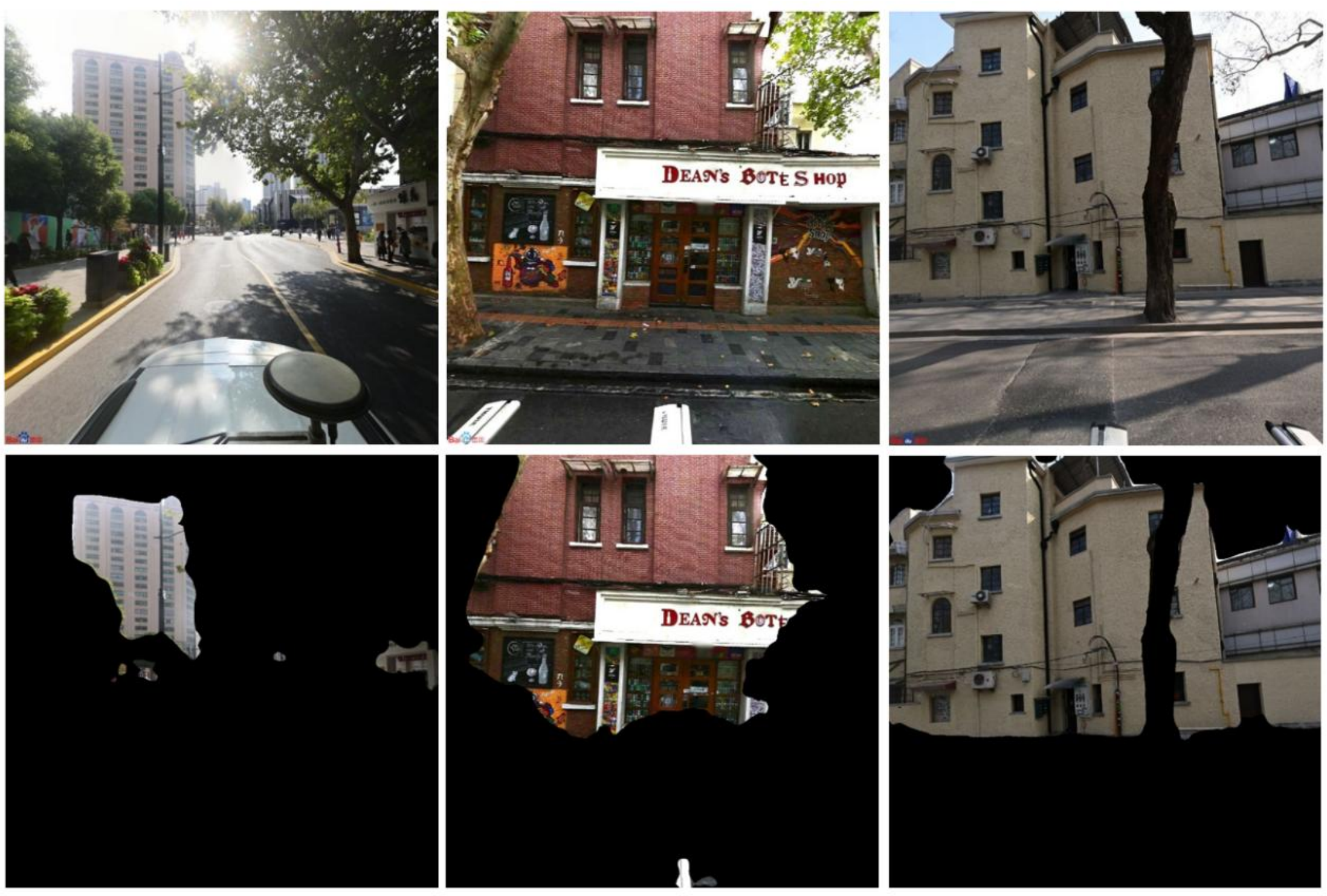

**Fig. 11** Building façade segmentation results before and after processing

On the other hand, Wujiang Road and West Nanjing Road—two street quarters within the same historical and cultural block—have a KS divergence of only 0.023. These areas are located less than 50 meters apart, which likely contributes to their similar visual characteristics. A similar pattern is observed between Xinhua Road and Yuyuan Road, both of which are known for their artistic atmosphere and creative industries. Their similar functions may be reflected in their shared color distribution. Despite localized differences, the dominant tones of building façade across most quarters are variations of beige and ochre. This suggests that, at a citywide level, the architectural color palette retains a degree of visual consistency, contributing to a unified urban identity.

To further examine the differences in building color composition between social media photos and real-world street views, we applied semantic segmentation to isolate building façade. This step involved removing indoor scenes, road surfaces, vehicles, sky, and other irrelevant elements to ensure a direct comparison of architectural color. Examples of the segmentation process are shown in Figure 11, which illustrate the contrast before and after isolating building façade.

When comparing the hue distributions of façades extracted from social media photos and street view images, we observed noticeable deviations. As shown in Figure 12, cool hues (e.g., green around H = 60 and blue around H = 120) are more prominent in social media content, whereas street-view baselines contain a higher proportion of warm tones, especially in the orange-yellow range (H = 10–30). This divergence can plausibly arise from multiple mechanisms rather than a single source. First, platform-side processing (e.g., compression and color-space conversion) and device-level computational photography (e.g., auto white balance and HDR) may alter color rendition. Second, capture conditions such as illumination, weather, time of day, and reflections from glass or water can systematically change perceived façade color. Third, user-side curation, including filters and manual adjustments, may shift color balance toward preferred styles. Because our dataset lacks standardized EXIF metadata and platform-side transformation logs, we do not attempt to attribute the observed shift to one dominant cause.

Nevertheless, the directionally consistent enrichment of cool hues across quarters suggests that the social media representation is not a neutral recording of on-site chromatic conditions. This pattern is consistent with documented tendencies for users to curate images to match platform aesthetics, which can increase cool-tone saturation or reduce warm casts to achieve a brighter, cooler look. We therefore interpret the façade hue shift as a perception–representation bias produced by combined technical and behavioral factors, and we treat the street-view distribution as a baseline for diagnosing this gap rather than as a ground-truth correction.

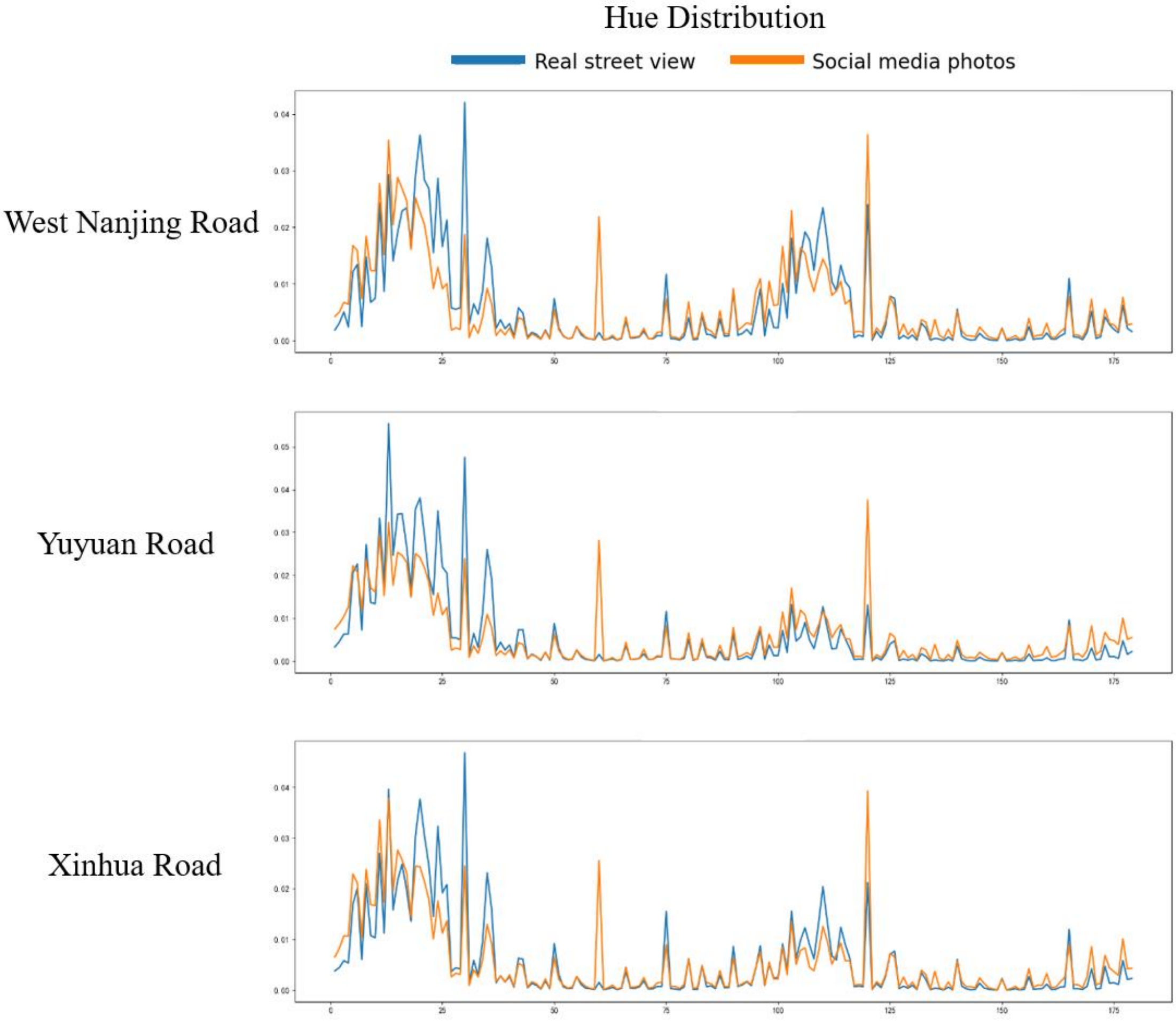

**Fig. 12** Comparison of color distributions between social media tourist photos and real-world street views for the West Nanjing Road, Yuyuan Road, and Xinhua Road quarters

### 3.3 Multidimensional Satisfaction Assessment

To qualitatively evaluate the multi-task BERT model's performance, we selected several representative tourist reviews and compared the predicted sentiment scores across the four dimensions with manual interpretations. As shown in Table 1, the model provides sentiment predictions ranging from −1 (negative) to 1 (positive) across four dimensions: Tourist Activities, Built Environment, Service Facilities, and Business Formats. These predictions are then examined in Table 2, which presents rational explanations to assess their consistency with human judgment.

Overall, the multi-task BERT model demonstrates strong capability in capturing sentiment across four dimensions: Tourist Activities, Built Environment, Service Facilities, and Business Formats. The predictions are largely consistent with human interpretation, especially in identifying dissatisfaction with commercial composition and praise for specific event-based activities.

**Table 1** Predicted sentiment scores for selected tourist reviews across four dimensions

| No. | Review Text (Origin) | Review Text (Translated) | Tourist Activities | Built Environment | Service Facilities | Business Formats |
|---|---|---|---|---|---|---|
| **1** | "今天去了武康大楼，真的很无聊，就搞不懂一栋楼怎么那么多的人，就可能很无聊吧，毕竟在上海打卡地太多了，就旁边的风景还可以，其他的就自行体会，旁边有一个名人故居可以去看看，去散散心也可以，就反正也还行吧" | "I went to Wukang Mansion today. Honestly, it was quite boring. I really don't get why so many people are drawn to just a building. Maybe it's just not that interesting. After all, there are so many other popular spots in Shanghai. The scenery nearby was okay, I guess, but the rest is up to personal experience. There's also a celebrity residence nearby worth visiting if you want a peaceful stroll." | **1** | **-1** | **0** | **-1** |
| **2** | "一直没有时间来上海打卡只是刷视频刷到过，看着很不错的样子，过年放假跟男朋友来打卡啦。真的很热闹而且人特别多哦，各个地方人都是来来往往挤的不行，但是玩的地方真的很多，外滩人也特别多，上厕所都要排很长的队哦，下次继续打卡" | "I never had time to visit Shanghai before, only saw it in videos. It looked great, so I finally came here during the New Year holiday with my boyfriend. It was really lively, and there were so many people everywhere—totally packed. But there were also plenty of places to have fun. The Bund was especially crowded, and even the restroom lines were super long. I'll definitely come back again." | **1** | **1** | **-1** | **0** |
| **3** | "很商业化了，没什么意思里面很小，没什么可逛的挺没必要花门票进去的。" | "It's way too commercialized and not interesting at all. The place is very small, and there's nothing worth seeing. Honestly, it's not worth the ticket." | **-1** | **-1** | **0** | **-1** |
| **4** | "到处是人，今年的灯会还是不错的。小吃价格太贵，关键好不好吃，建议不要在豫园里消费。" | "People everywhere. But this year's lantern festival was actually quite nice. The snacks were way too expensive though—and to be honest, not even that tasty. I wouldn't recommend buying anything inside Yuyuan Garden." | **1** | **1** | **0** | **-1** |

**Table 2** Explanation of prediction rationality for each dimension

| No. | Dimension | Score | Explanation of Prediction Rationality |
|---|---|---|---|
| **1** | Tourist Activities | **1** | The reviewer explicitly states it was "quite boring" and that there was nothing much to do, indicating low activity appeal. The positive score is thus unreasonable. |
| **1** | Built Environment | **-1** | Complaints about overcrowding ("so many people") and a dismissive comment about the building suggest a negative perception of the physical environment. The score is reasonable. |
| **1** | Service Facilities | **0** | No mention of specific facilities (e.g., restrooms, seating). The neutral score is reasonable due to insufficient information. |
| **1** | Business Formats | **-1** | The reference to "so many other popular spots in Shanghai" implies fatigue or criticism of homogeneity and commercialization. The score is reasonable. |
| **2** | Tourist Activities | **1** | The reviewer praises the abundance of fun activities ("plenty of places to have fun"), indicating high satisfaction. The score is reasonable. |
| **2** | Built Environment | **1** | Although the environment "looked great," the reviewer also mentions severe crowding. A neutral rating may have been more appropriate, but the positive score is mostly reasonable. |
| **2** | Service Facilities | **-1** | "Long lines for the restroom" is a direct complaint about inadequate facilities. The score is reasonable. |
| **2** | Business Formats | **0** | No direct mention of business formats. The neutral score is reasonable. |
| **3** | Tourist Activities | **-1** | "Nothing worth seeing" directly reflects a lack of engaging activities. The score is reasonable. |
| **3** | Built Environment | **-1** | The phrase "very small" criticizes cramped space, implying poor environmental experience. The score is reasonable. |
| **3** | Service Facilities | **0** | The comment about the ticket price may imply dissatisfaction with value-for-money, but no explicit facility issue is raised. The neutral score is somewhat reasonable. |
| **3** | Business Formats | **-1** | "Too commercialized" suggests a negative view of homogeneous or overwhelming commercial presence. The score is reasonable. |
| **4** | Tourist Activities | **1** | The lantern festival is described positively ("quite nice"), indicating strong activity quality. The score is reasonable. |
| **4** | Built Environment | **1** | The review does not directly mention the built environment. The positive score is therefore unreasonable. |
| **4** | Service Facilities | **0** | No specific mention of facilities like restrooms or seating. The neutral score is reasonable due to lack of information. |
| **4** | Business Formats | **-1** | Complaints about overpriced and low-quality food, with a suggestion not to spend inside, reflect dissatisfaction with commercial offerings. The score is reasonable. |

# 4 Discussion

This study operationalizes tourist perception in historic urban quarters through a multimodal framework that links visual attention, chromatic representation, and dimension-specific satisfaction. By combining semantic segmentation, color palette comparison, and multi-task sentiment modeling, we move beyond treating images and texts as separate outputs and instead trace how foregrounded streetscape elements and curated color themes relate to narratives of praise and complaint. In doing so, the framework responds to calls in tourism and heritage research to connect the cognitive and affective dimensions of destination image with observable traces of the tourist gaze in urban space.

Compared with prior multimodal studies that either visualize heritage elements from social media photos or use deep learning to classify tourist images without explicit links to experience components, our approach emphasizes integration and interpretability. For example, work such as Bai et al. focuses on mapping perceived heritage features, while other studies use geo-tagged photos to infer destination image or spatiotemporal behavior from visual content and reviews considered in isolation. By contrast, we explicitly connect three stages of perception: what is visually foregrounded in shared photos, how these representations deviate from street-view baselines, and how visitors evaluate activities, built settings, services, and commercial offerings in text. This analytic chain helps clarify why certain mismatches between represented and on-site environments matter for satisfaction, rather than treating visual and textual signals as parallel but unrelated indicators.

The findings also speak to longstanding discussions about the management of the "tourist-historic city". In several leisure-oriented quarters, greenery and heritage buildings occupy a large share of the visual field, consistent with evidence that vegetated streetscapes and historic fabric can enhance preference, comfort, and place attachment. At the same time, our color analysis shows that tourists often post images with cooler, highly curated palettes that diverge from warmer street-level façades, suggesting that online representations emphasize particular atmospheric qualities rather than faithfully recording on-site conditions. Dimension-specific sentiment further reveals where dissatisfaction clusters around overcrowding, service facilities, or commercial composition, even when the built environment is evaluated more positively. Taken together, these patterns highlight that conservation policies and design guidelines cannot focus solely on architectural control; they also need to address experiential qualities, such as walkability, greenery, signage, and the balance between everyday uses and tourism-oriented commerce, in line with heritage management and place-making traditions.

From an applied perspective, the proposed indicators can support visitor-oriented urban design and heritage governance in several ways. Pixel-level metrics on visual focus and greenery can inform streetscape maintenance and greening strategies, while perception–reality gaps in color composition can signal where façade renovation or lighting schemes fail to match what visitors expect or value. The four satisfaction dimensions translate unstructured reviews into a compact diagnosis of where interventions are most needed, whether in programming activities, improving service facilities such as restrooms and seating, or regulating business formats to avoid perceived over-commercialization. Because these dimensions echo established structures in servicescape and heritage tourism research, they can be interpreted alongside survey-based evidence rather than standing apart from existing planning practice.

Methodologically, each module can still be strengthened. For semantic segmentation, we adopt DeepLabV3+ with a MobileNetV2 backbone to balance accuracy and efficiency, yet fine-grained objects and visually complex scenes remain challenging. Future work can explore stronger segmentation architectures and improve robustness through more diverse training data and data augmentation. For satisfaction prediction, our current analysis relies on textual reviews, even though visitor evaluations are shaped by the interplay between experiences and physical settings. Integrating street-level visual features such as façade condition, greenery, perceived cleanliness, and crowding into sentiment models would provide additional contextual grounding and improve interpretability. For color analytics, social media photos are not neutral recordings: platform-specific compression, filters, and device-level processing can systematically alter color rendition. Comparative studies across platforms and cultures, along with learning-based calibration strategies that align social media color distributions with street-view baselines, could reduce bias in palette comparisons and help distinguish aesthetic preference from technical transformation.

This study is also subject to several study-level limitations. First, user-generated content is not a representative sample of all visitors; posting behavior varies by demographics, trip purpose, and platform norms, which can bias both visual and textual signals. Second, photos on Dianping are captured with heterogeneous devices, including smartphones in automatic modes and digital cameras with manually tuned parameters. We cannot fully standardize exposure and color processing across these sources. Although we apply white balance correction and use street views as a baseline, residual distortions related to device and capture settings are likely to remain, especially for fine-grained palette comparisons. Third, temporal factors such as season, time of day, and weather influence both photography behavior and perceived color, and the current analysis does not explicitly disentangle these effects. Fourth, the sentiment dataset relies on automated labeling using a large language model, which may introduce noise; future work should incorporate larger human-labeled benchmarks and uncertainty-aware evaluation. Finally, the Shanghai case provides an exploratory validation. Transfer to other cities will require adapting the semantic taxonomy to local heritage typologies, recalibrating the satisfaction dimensions to reflect context-specific concerns, and examining whether perception–reality relationships differ across visitor groups and cultural backgrounds.

## 5 Conclusion

This study develops and applies an AI-powered, multimodal framework to decode tourist perception in historic urban quarters using geo-tagged photos and review texts from a major Chinese platform, complemented by street-view imagery. Conceptually, the framework treats perception as a linked process that spans attention, representation, and appraisal. Semantic segmentation quantifies which streetscape elements are foregrounded in tourist-shared images, color palette comparison reveals how online representations differ from on-site façades, and multi-task sentiment modeling captures how visitors evaluate activities, built settings, service facilities, and commercial offerings. Together, these modules address methodological fragmentation in user-generated content research and provide an interpretable bridge between deep learning outputs and established constructs in destination image, servicescape, and heritage tourism.

Empirically, the Shanghai case demonstrates that tourist photos consistently highlight key components of historic quarters, including greenery, heritage buildings, artistic installations, and food-related scenes, while color distributions in social media content often shift toward cooler, curated palettes that depart from warmer street-level façades. These patterns suggest that online imagery reflects not only what is present on-site but also what visitors wish to project about the atmosphere of a place. Dimension-specific sentiment analysis further shows how praise and criticism are distributed across activities, physical settings, supporting services, and commercial formats, offering a nuanced diagnosis that can guide heritage management and visitor-oriented urban design.

The framework is designed to be transferable. With appropriate adaptation of the semantic taxonomy and satisfaction dimensions, it can be applied to other historic quarters and urban contexts to compare perception–reality gaps across heritage typologies and visitor cultures. Future work can extend the approach by integrating additional data sources, such as mobility traces or planning documents, and by coupling perception indicators with behavioral or economic outcomes. More broadly, the study illustrates how AI techniques can be embedded within theory-informed analyses of urban experience, turning large-scale digital traces into planning-relevant evidence for sustaining both the cultural significance and everyday livability of historic urban quarters.

## Appendix A

### Appendix A.1 Semantic Segmentation Category List

- Human
- Door
- Wall
- Building
- Stairs
- Sign
- Advertisement
- Light
- Tree
- Flower
- Plant
- Animal
- Toy
- Food
- Drink
- Vehicle
- Bench
- Fence
- Fountain
- Statue
- Vendor
- Artwork

### Appendix A.2 Semantic Segmentation Hyperparameter Settings

- Initial Epoch: 0
- Frozen Training Epochs: 50
- Unfrozen Training Epochs: 400
- Batch Size (Frozen): 32
- Batch Size (Unfrozen): 16
- Optimizer: SGD

- Initial Learning Rate: 7e-3
- Minimum Learning Rate: 7e-5
- Momentum: 0.9
- Weight Decay: 1e-4
- Learning Rate Schedule: Cosine Annealing
- Mixed precision training (FP16) enabled

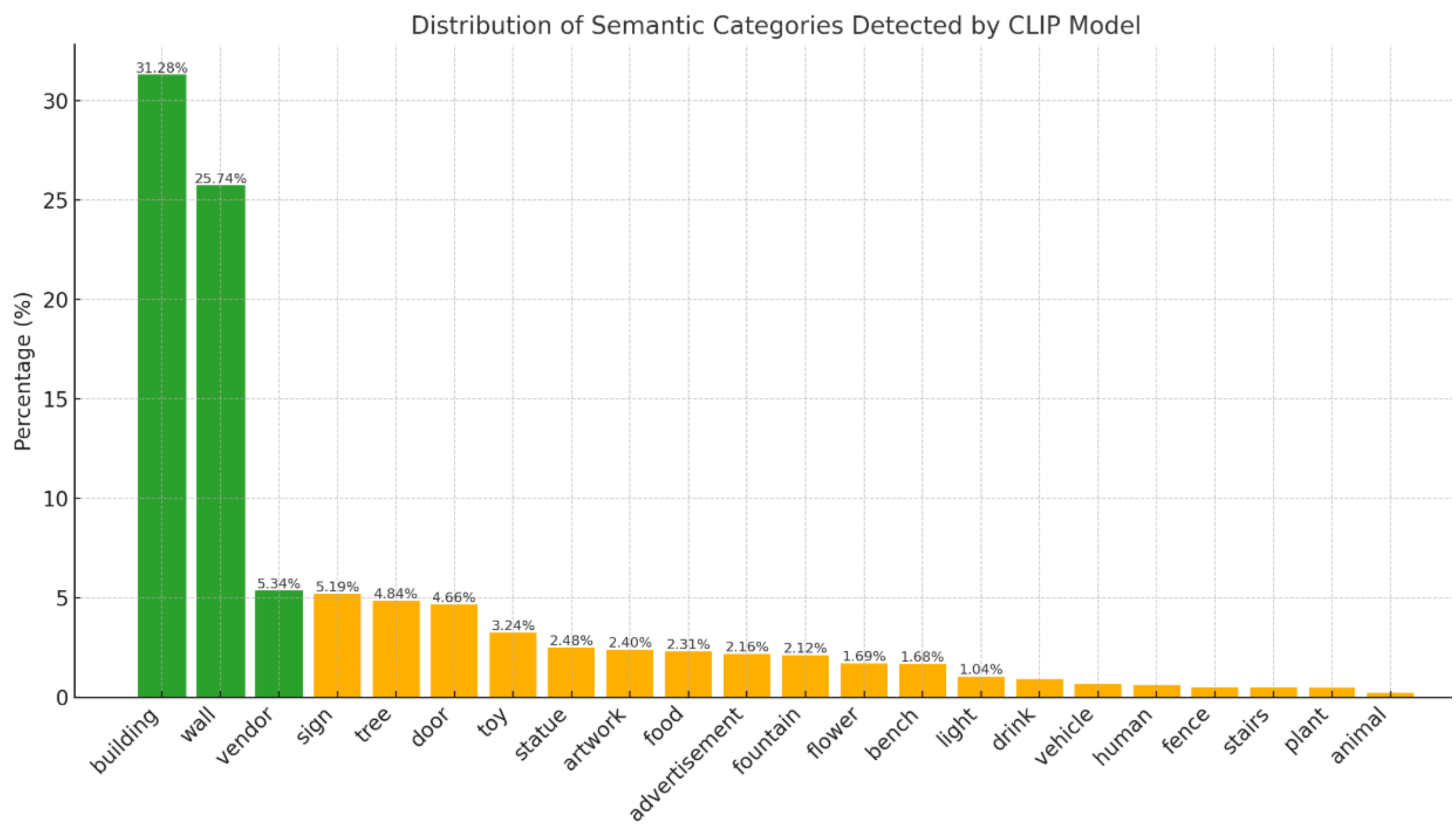

**Fig. A1** Distribution of semantic categories detected by CLIP Model

### Appendix A.3 Chinese Color System

In order to standardize color classification, the Chinese color system is adopted with the following structure:

- Five primary hues: Red (R), Yellow (Y), Green (G), Blue (B), Purple (P)
- Five intermediate hues between adjacent primaries: Red-Yellow (YR), Yellow-Green (GY), Green-Blue (BG), Blue-Purple (PB), Purple-Red (RP)
- Ten basic hues are formed from the primaries and intermediates
- Each pair of basic hues is further divided into four equal intervals, producing 40 hue levels (H)
- Saturation (S) and Brightness (V) are each divided into 5 levels, resulting in a total of $40 \times 5 \times 5 = 1000$ color categories

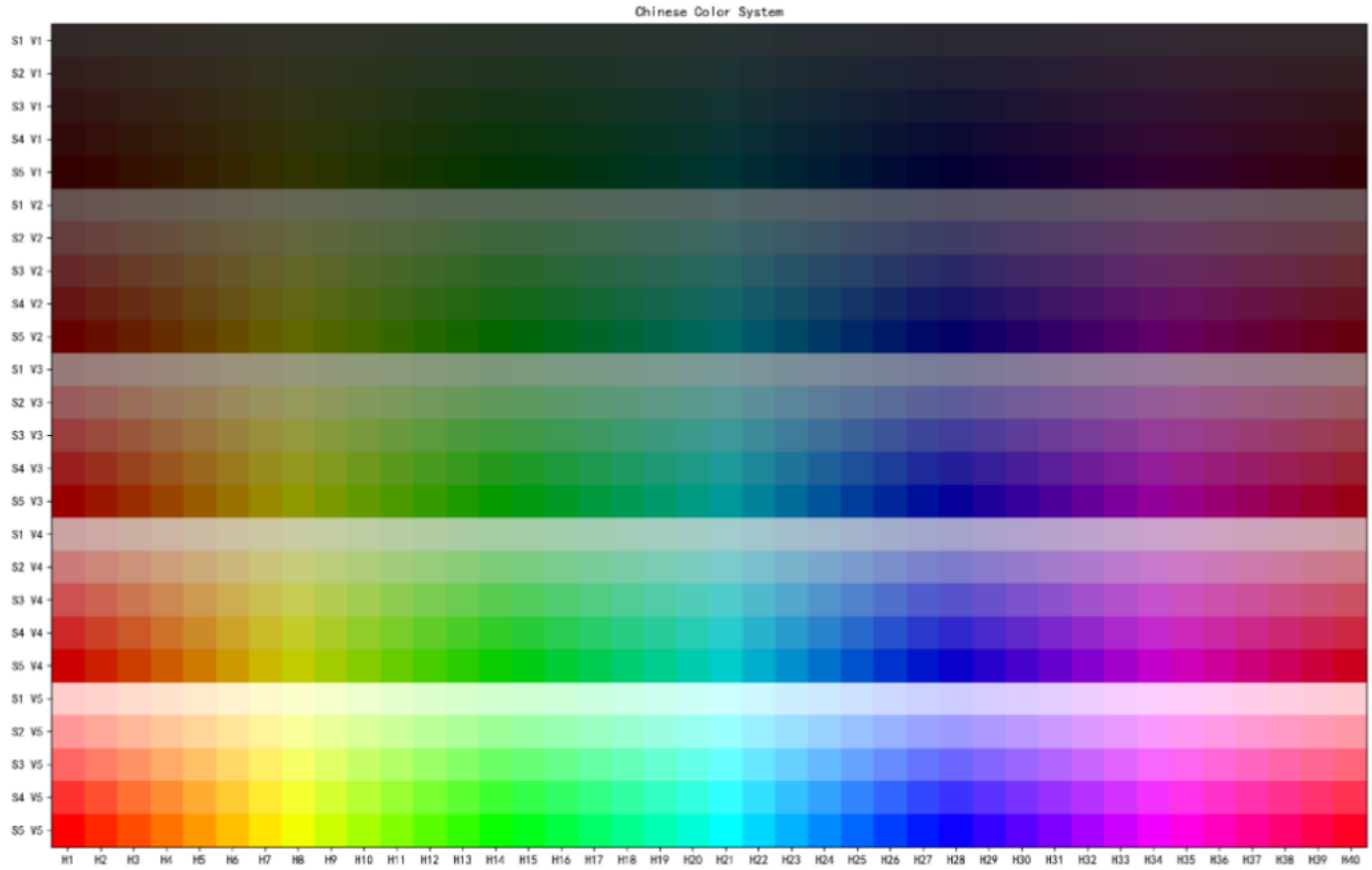

**Fig. A2** Structure of Chinese color system

### Appendix A.4 Prompt Design for Sentiment Annotation

Please rate the sentiment of the review content across the following four dimensions: [Activities, Built Environment, Service Facilities, Business Formats]

**[Dimension Definitions]**

1. **Activities**: Refers to spontaneous individual activities (e.g., dining, taking photos, sightseeing, shopping) or organized special events (e.g., fashion shows, exhibitions, markets).
2. **Built Environment**: Includes architectural landscapes, landmarks, street environment, natural elements, and small-scale urban features (e.g., aesthetic quality of buildings, street cleanliness, greenery).
3. **Service Facilities**: Refers to the provision of public services, transportation, and tourist assistance (e.g., buses, restrooms, seating areas, signage) as well as service attitude.
4. **Business Formats**: Reflects the diversity and quality of commercial offerings (e.g., food, jewelry, entertainment, fashion, accommodation, cultural/creative retail), perceived value, and shop uniqueness.

**[Scoring Rules]**

- 1 = Positive (satisfied), 0 = Neutral, -1 = Negative (dissatisfied)
- Return four numbers separated by spaces only
- **Examples**:
  - "Visited the pedestrian street—check! But honestly, the street was a bit dirty and lacked dining options." → **1 -1 0 -1**
  - "It was overcrowded and felt unsafe, but the view was stunning and the architecture was impressive." → **-1 1 0 0**

**Please rate the following review:**

"{text}"

**Appendix A.5 BERT-Based Model Configuration**

- Backbone: BERT-base-Chinese
- Tokenization: max_length=128, padding=True, truncation=True
- Label Mapping: {-1, 0, 1} → {0, 1, 2}
- Classifier: Four independent Linear(hidden_size, 3) layers with Dropout(0.3)
- Optimizer: AdamW, lr=1e-5, weight_decay=0.2, warmup_ratio=0.2
- Loss Function: CrossEntropyLoss (summed over four dimensions)
- Gradient Clipping: max_grad_norm=1.0
- Batch Size: 8 (train) / 32 (val)
- Epochs: 20
- Precision: Mixed precision (FP16)
- Evaluation Metric: macro_f1_score on each dimension

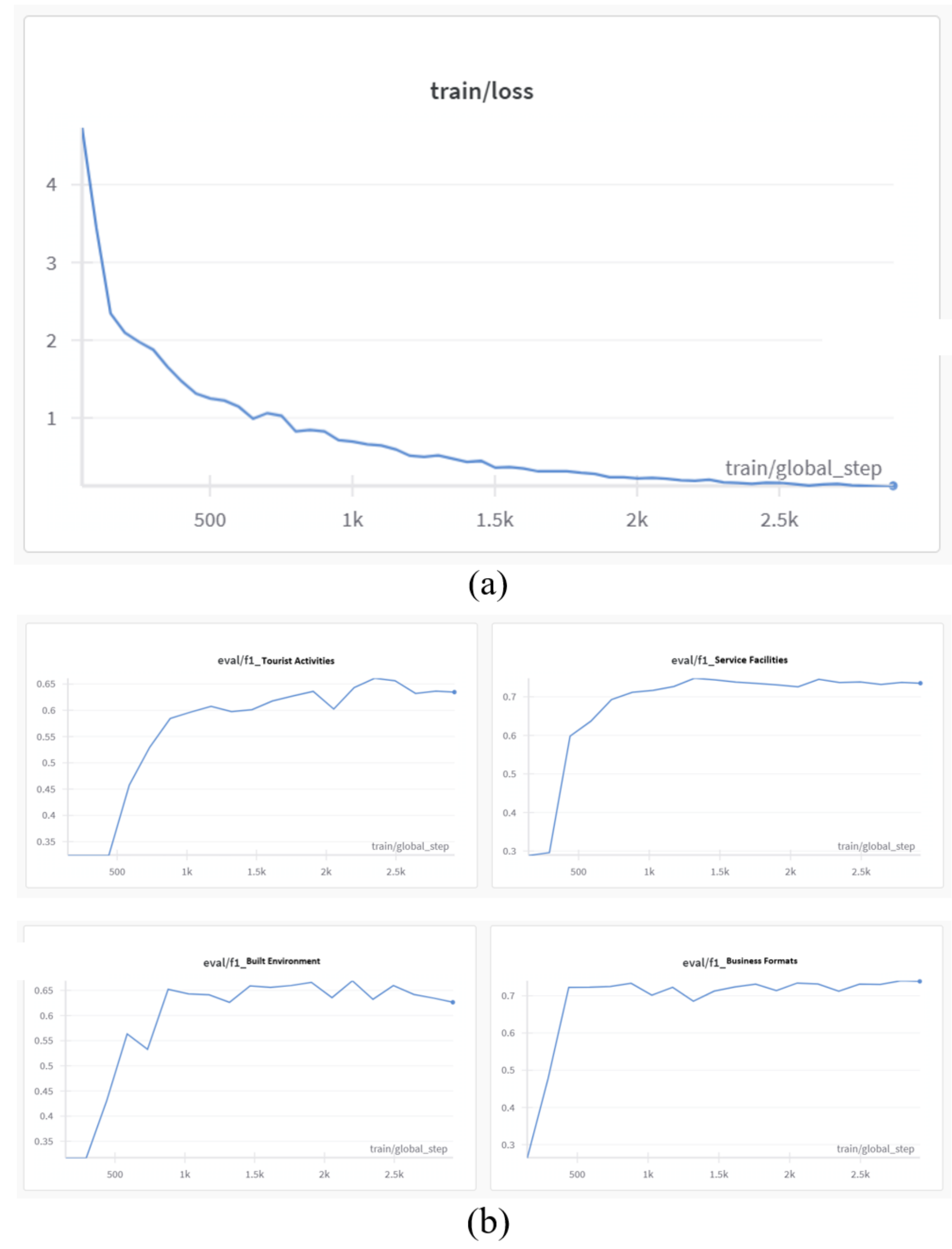

**Fig. A3** Multi-task BERT training loss (a) and evaluation F1 scores across 4 dimensions (b)

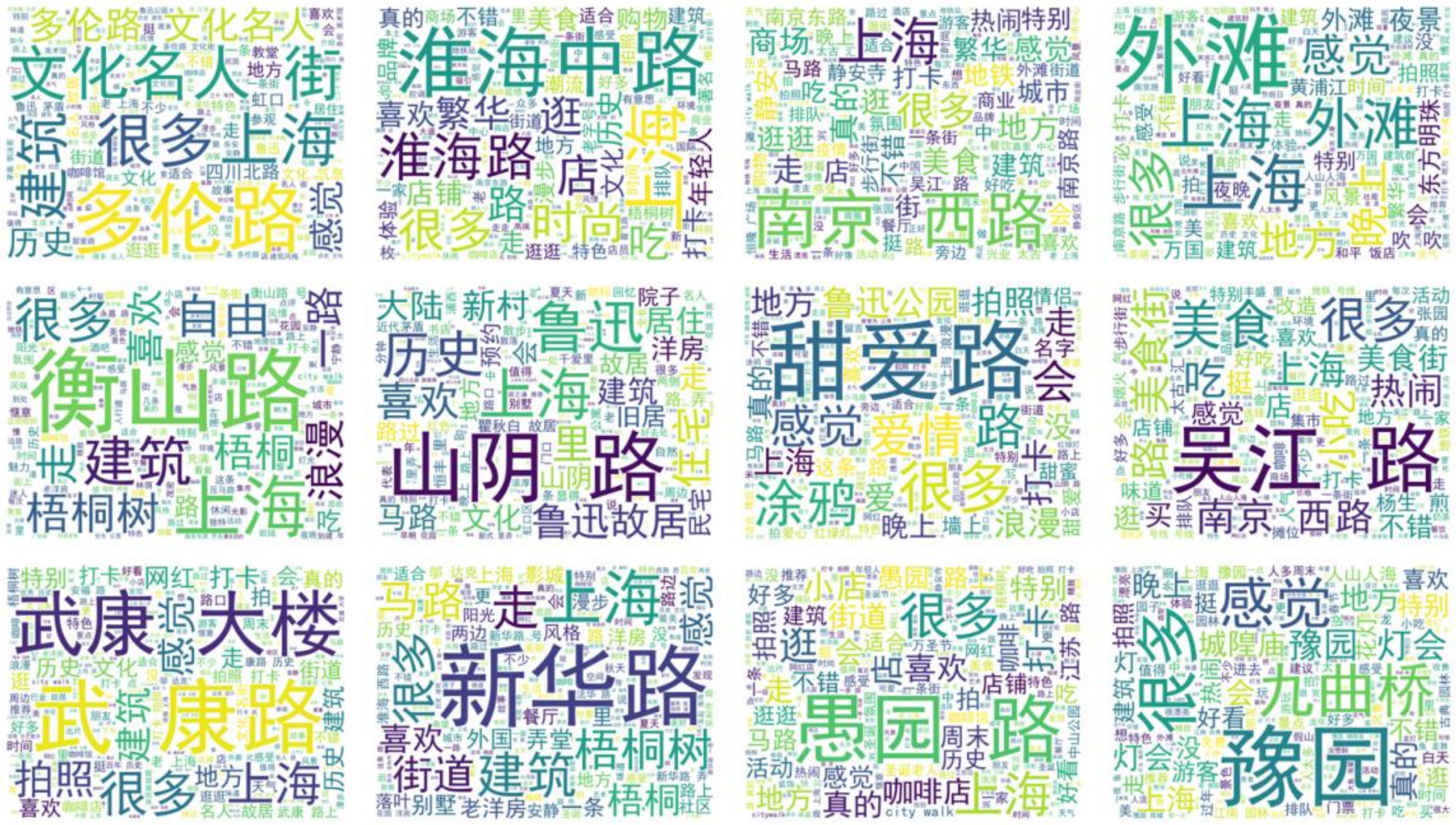

**Fig. A4** Chinese word clouds for the 12 historic urban quarters in Shanghai